\documentclass[journal]{IEEEtran}
\ifCLASSOPTIONcompsoc
\usepackage[nocompress]{cite}
\else
\usepackage{cite}
\fi
\ifCLASSINFOpdf
\else
\fi
\usepackage{multirow,palatino, epsfig, latexsym, algorithmic, algorithm, epstopdf, amsmath, amssymb, color, amsthm,graphicx,enumitem, hyperref}

\newcommand{\ignore}[1]{}

\DeclareMathOperator*{\argmax}{argmax}

\newtheorem{myRemark}{Remark}

\newtheorem{myDefinition}{Definition}

\begin{document}
\title{Optimal-margin evolutionary classifier}

\author{Mohammad Reza Bonyadi, David C. Reutens
		
\thanks{All authors are with the Centre for Advanced Imaging (CAI), the University of Queensland, Brisbane, QLD 4072, Australia. M. R. Bonyadi (reza@cai.uq.edu.au, rezabny@gmail.com) is also with the Optimisation and Logistics Group, The University of Adelaide, Adelaide 5005, Australia.}}

	
\IEEEtitleabstractindextext{%
\begin{abstract}
We introduce a novel approach for discriminative classification using evolutionary algorithms. We first propose an algorithm to optimize the total loss value using a modified 0-1 loss function in a one-dimensional space for classification. We then extend this algorithm for multi-dimensional classification using an evolutionary algorithm. The proposed evolutionary algorithm aims to find a hyperplane which best classifies instances while minimizes the classification risk. We test particle swarm optimization, evolutionary strategy, and covariance matrix adaptation evolutionary strategy for optimization purpose. Finally, we compare our results with well-established and state-of-the-art classification algorithms, for both binary and multi-class classification, on 19 benchmark classification problems, with and without noise and outliers. Results show that the performance of the proposed algorithm is significantly (t-test) better than all other methods in almost all problems tested. We also show that the proposed algorithm is significantly more robust against noise and outlaiers comparing to other methods. The running time of the algorithm is within a reasonable range for the solution of real-world classification problems. 
\end{abstract}
		
}
\maketitle
\IEEEdisplaynontitleabstractindextext
\IEEEpeerreviewmaketitle

\section{Introduction}
\label{sec:intro}
\IEEEPARstart{T}{he} main goal of a supervised classification algorithm is to identify the class to which each instance belongs based on a given set of correctly labeled instances. There are two types of classifiers, generative and discriminative \cite{ng2002discriminative}. While generative classifiers learn a joint distribution between inputs and class labels, discriminative classifiers learn the exact class label from the training dataset. It has been shown \cite{ng2002discriminative} that "discriminative classifiers are almost always to be preferred".

A discriminative classifier \cite{ng2002discriminative} is defined as follows:
\begin{myDefinition}
(Discriminative classifier)
Let $ S,S_{1},S_{2},...,S_{c} $ be sets of instances such that, $ \forall i,j \in \{0,...,c-1\} $, $ i \ne j $, $ |S_i|=m_i $, $ |S|=m $, $ S_{i} \cap S_{j} = \emptyset $, and $ \cup_{i=0}^{c-1}S_i=S $. A classifier $\psi_\beta:S \to Y$, $ Y=\{0,...,c-1\} $, aims to guarantee
\begin{equation*}
	\forall i \in Y, \forall x \in S_i, P(\psi_\beta(x)=i|x)=1,
\end{equation*}	
\noindent where $\beta$ is a set of configurations for the procedure $\psi_\beta(x)$, and $P$ is the probability measure.\footnote{Throughout this paper, we consider a special case of classification problems where all members of $ S $ are in $ \mathbb{R}^n $ (so called feature space), and each instance in $ S $ is represented by a vector. We also assume that feasible values for $ x_j $ (called a variable throughout this paper), the $ j^{th} $ element of the instance $ \vec{x} $, are ordered by the operator "$ \le $" (i.e. $ x_j $ is not categorical). }
\end{myDefinition}

The classifier $ \psi_\beta(\cdot) $ is usually a combination of an optimization problem, $ \Omega_\beta $, a transformation $ \mathcal{M}_\beta: S \to \hat{S} $, and a discriminator $ \mathcal{D}: \hat{S} \to Y $. The solution to $ \Omega_\beta $ yields $ \beta $ that transforms any given instance $ x $ to $ \hat{x} $ through $ \hat{x} = \mathcal{M}_\beta(x) $, which is finally mapped into the class label by $ \mathcal{D}(\hat{x}) $. In reality, the true class of only a subset of $ S $ is known (the training set). It is hence challenging to find a transformation $\beta$ for which $\psi_\beta(\vec{x})$ is the true class of all $ \vec{x} $ in $ S $, including the ones that are not in the training set (unseen instances). Therefore, the generality of an optimized $ \beta $ depends on the assumptions made to formulate $ \Omega_\beta $ and the given training set itself. We assume that $ \mathcal{M}_\beta(\vec{x}) $ is linear ($ \beta = <\omega, b> $ and $ \mathcal{M}_\beta(\vec{x})=\vec{x}\omega^T+b $). For binary classification ($ y_i \in \{-1, 1\} $ for all $ i $), $ \omega: \mathbb{R}^n \to \mathbb{R} $ and $ b \in \mathbb{R} $ represent the normal vector and the intercept of a hyperplane that separates the two classes, the \textit{separator hyperplane}. Note that non-linear classification is possible by adding a kernel to a linear classifier \cite{haykin2004comprehensive}. Also, the algorithms for binary classification can be extended to work with multi-class classification  through the one-vs-one approach \cite{hsu2002comparison}.

\textbf{Motivation}: One way to formulate the optimization problem $ \Omega_\beta $ for discriminative classification is through minimization of the number of misclassified instances, i.e., minimization of the total loss of the 0-1 loss function \cite{nguyen2013algorithms} (see Section \ref{sec:lossfunction}). Optimization of the total loss over the 0-1 loss function leads to finding the optimal separator hyperplane that is robust against outliers \cite{nguyen2013algorithms}. As this optimization problem is NP-hard \cite{ben2003difficulty}, existing algorithms to find such a hyperplane are impractical for real-world classification \cite{nguyen2013algorithms}. In addition, the original definition of the 0-1 loss function leads to inefficiencies such as sensitivity to class imbalance and ignorance of classification risk. Therefore, several alternative formulations have been proposed \cite{cortes1995support,vapnik1998statistical} that turn this formulation into a smooth convex one and use gradient descent to solve it efficiently. These alternative formulations, while convex and can be solved to global optimum, are sensitive to noise and outliers. 

In this paper, we propose an alternative loss function to address the shortcomings associated with the 0-1 loss function (sensitivity to imbalances in the number of instances in each class and ignorance of classification risk) while keep its advantages (robustness against noise and outliers). We then propose an efficient algorithm to find the optimal solution to the one-dimensional discriminative classification problem using this loss function. We extend that algorithm to handle multi-dimensional classification problems using an evolutionary algorithm. Three evolutionary algorithms (particle swarm optimization, evolutionary strategy, and covariance matrix adaptation evolutionary strategy) are tested and compared for this purpose. We finally extend the algorithm to work with multi-class classification problems. Note that evolutionary algorithms have been used for generative classification \cite{jia2016optimized,ding2013evolutionary} (e.g., training of neural networks weights and/or structures). Under some conditions, however, the generative approaches are not optimal for discriminative classification tasks \cite{ng2002discriminative,vapnik1998statistical}, that motivates our proposal for a better discriminative evolutionary-based classifier.

We structure the paper as follows: Section \ref{sec:background} provides a background on the classification methods used for comparison. Section \ref{sec:proposed} details our proposed method. Section \ref{sec:experiment} reports and discusses the results of the comparisons between our method and 6 other classification methods based on 19 standard benchmark classification problems. Section \ref{sec:conclusions} concludes the paper and points to potential future directions.

\section{Background}
\label{sec:background}
This section provides background information on loss functions, optimization of the 0-1 loss function, and state-of-the-art classification methods.

\subsection{Loss function and total loss}
\label{sec:lossfunction}
\begin{figure}
	\centering
	\includegraphics[width=0.49\textwidth]{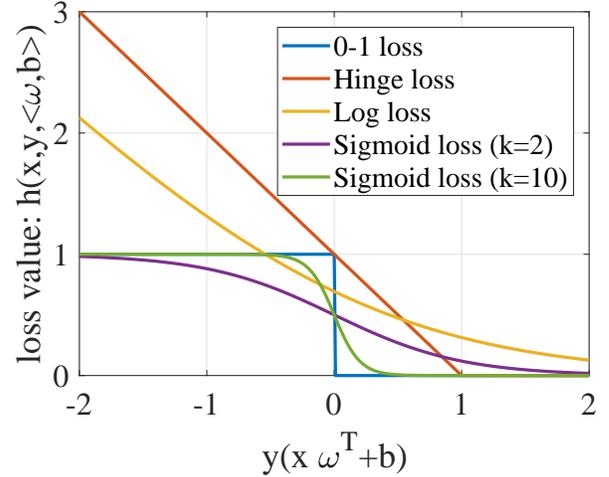} 
	\caption{Loss functions frequently used for classification.}
	\label{fig:losses}
\end{figure}

One of the most frequently-used definitions for the optimization problem $ \Omega_\beta $ is based on "total loss" and "loss function" ($ \beta=<\omega, b> $): 
\begin{equation}\label{eq:general-loss}
\Omega_\beta: \min \limits_{\beta} {\sum_{i=1}^{m}h(y_i,x_i,\beta)} + \alpha\mathcal{R}(\omega)
\end{equation}
\noindent where $ h(.) $ is a \textit{loss function} that measures the difference between the class label $ y_i $ and the class label calculated by the classification algorithm ($ \mathcal{D}(\mathcal{M}_\beta(\vec{x}_i)) $), and $ {\sum_{i=1}^{m}h(y_i,x_i,\beta)} $ is the \textit{total loss}. The function $ \mathcal{R}: \mathbb{R}^n \to \mathbb{R} $ is the \textit{regularization function} and $ \alpha $ is the \textit{regularization factor}. The aim is then to minimize the total loss, given the regularization factor $ \alpha $. The role of regularization is to control the balance between the bias and the variance of the model \cite{bishop2006pattern}. Frequently-used regularization functions with this settings are $ L_1 $ ($ \mathcal{R}(\omega)=||\omega||_1 $, also known as least absolute shrinkage and selection operator, LASSO, \cite{tibshirani1996regression}) and $ L_2 $ ($ \mathcal{R}(\omega)=||\omega||_2 $, also known as Tikhonov regularization \cite{bishop2006pattern}). Given $ \beta = <\omega, b> $, the simplest loss function is the \textit{0-1 loss function} that generates a '1' for a misclassified instance and a '0' for a correctly classified instance.
\begin{equation}
\label{eq:0-1-loss}
h(y,x,\beta)=
\begin{cases}
1 & y(x\omega^T+b) \le 0 \\
0 & y(x\omega^T+b)>0
\end{cases}
\end{equation}
If $ y_i $ and $ \vec{x}_i \omega^T+b $ have the same sign then the instance $ \vec{x}_i $ has been classified correctly using $ \omega $ and $ b $. The optimal $ \omega $ and $ b $ lead to the minimum total loss, i.e., minimum number of misclassified instances in $ X $ (number of instances $ \vec{x}_i $ for which $ y_i(\vec{x}_i \omega^T +b)<0 $). The best solution to this optimization problem provides the normal vector ($ \omega $) and the intercept ($ b $) of the optimal separator hyperplane. This problem is, however, NP-hard \cite{ben2003difficulty} mainly because of the definition of $ h(.) $. Hence, many studies have proposed other differentiable smooth variations of this loss function so that the final optimization problem is solvable using a gradient descent algorithm. For example, the hinge loss, defined by $h(y_i, \vec{x}_i, <\omega, b>)= max(0,1-y_i(\vec{x}_i \omega^T +b)) $, has been used in support vector machines \cite{cortes1995support} and the log loss, defined by $ h(y_i, \vec{x}_i, <\omega, b>)=\log(1+e^{-y_i(\vec{x}_i \omega^T +b)}) $ has been used in the logistic regression \cite{vapnik1998statistical}. An alternative to the log loss is the sigmoid loss, $ \frac{1}{1+e^{-ky(\vec{x}\omega^T+b)}} $, where $ k $ is a constant (Fig. \ref{fig:losses} shows some well-known loss functions). 

Using the 0-1 loss function to calculate the total loss leads to identification of a hyperplane that, unlike its smooth alternatives, is known to be robust to outliers \cite{nguyen2013algorithms}. The reason is that the loss value does not scale with the misclassified instances in the 0-1 loss function while it does scale with them in hinge and log loss, i.e., the smaller the value of $ y_i(\vec{x}_i \omega^T +b) $, the larger the impact on the total loss. 

\subsection{Direct optimization of 0-1 loss function}
Direct optimization of the total loss with the 0-1 loss function, while leading to an optimal separator hyperplane, is an NP-hard problem. In \cite{nguyen2013algorithms}, the authors proposed multiple algorithms to solve this problem to optimality and near optimality. For example, they used the branch and bound method, in combination with linear programming, to solve the problem to optimality. The time complexity of the method, however, was exponential, making it impractical for real-world classification. Another method proposed was based on the fact that an optimal hyperplane can be built upon $ n $ instances ($ n $ is the number of dimensions) from the training set. Hence, they proposed that all possible combinations of $ n $ instances be tested to find the combination upon which the best separator hyperplane can be built. The time complexity of this method is also exponential with the number of instances and dimensions. The fastest efficient method proposed was based on optimization of the sigmoid loss with variable $ k $. As larger values for the $ k $ leads to a better approximation of the 0-1 loss function, the sigmoid loss with larger $ k $ would lead to convergence to the optimal hyperplane. However, larger values for $ k $ cause the search space to become non-convex, making it difficult to solve the problem to optimality. Hence, it was proposed that $ k $ be increased from 2 to 200 and the total loss function optimized locally for each $ k $, given the solution found for the previous value of $ k $. The method, called Smooth Logistic Algorithm (SLA), was shown to effectively solve binary classification problems to near optimality with a much better time complexity compared to the branch and bound and point selection methods. Although SLA outperformed support vector machines and logistic regression, its running time was still impractical even for a small number of dimensions.

\subsection{Classification approaches with smoothed 0-1 loss function}
We describe in brief popular classification methods used herein for benchmarking and comparison.

\subsubsection{Support vector machine (SVM) and its recent extensions}
SVM aims to find the normal vector ($ \omega $) and the intercept ($ b $) of the separator hyperplane such that the distance between the closest instances from each class (i.e., support vectors) to the hyperplane is maximized (margin) \cite{suykens1999least} while the instances are classified correctly. The balance between the separation accuracy and the margin is adjusted by a variable $ \lambda $. SVM can be formulated by a hinge loss plus the margin term and optimized by a gradient descent. The separation is determined by the sign of $ \vec{x}\omega^T + b $ ($ T $ is transpose) indicating the side of the hyperplane to which the instance, $ \vec{x} $, belongs. Note that this approach prioritizes hyperplanes that provide minimum empirical risk (maximum margin) measured by the inverse of the distance between the hyperplane and the closest instances from each class. This risk is of utmost importance especially in classification of noisy datasets \cite{suykens1999least}.

Twin SVM (TSV) \cite{shao2011improvements,khemchandani2007twin} seeks a pair of hyperplanes, one of which is closer to the instances from class $ -1 $ than instances from class 1 and the other is closer to the instances from class 1 compared to the instances from class $ -1 $. The discriminator is then used to calculate which hyperplane is closer to an instance. A recent extension \cite{shao2015weighted} of TSV is based on the introduction of a weighted linear loss to the formulation of TSV (and hence is called WSV throughout the paper) instead of the hinge loss, reducing the quadratic problem to a linear one.

\subsubsection{Minimax probability machine (MPM) and its recent extensions}
Minimax probability machine (MPM) is a type of discriminative classifier that aims to minimize the maximum misclassification probability of instances \cite{lanckriet2002robust}. MPM attempts to find a generalizable margin by paying attention to the distributions within classes rather than the instances themselves. Evidence \cite{xue2011structural} suggests that the structure of the instances in different classes provides important information for the design of generalizable transformations for classification. Structural minimax probability machine (SMP) \cite{gu2017structural} makes use of the structural information, approximated by two finite mixture models in each class, in the context of MPM for classification of instances. This idea has been shown to be very effective on a set of standard datasets. 

\subsubsection{Logistic regression (LR) and its recent extensions}
Logistic regression (LR) \cite{walker1967estimation} provides a smooth estimation of the 0-1 loss function using the sigmoid function (see Fig. \ref{fig:losses}), usually with $ k=2 $. Elastic-net extends LR by incorporating the LASSO and Tikhonov regularization terms. For classification purposes, elastic net usually performs optimization in relation to the true class labels, restricting the algorithm to binary targets. This restriction was resolved in \cite{zhang2017discriminative} (Discriminative Elastic Least-square, DEL) where a term was used to relax the class labels. The optimization problem associated with this classification process was then introduced and solved via an iterative procedure. 

\subsubsection{Linear discriminant analysis (LDA) and its recent extensions}
A geometrical interpretation for $ \omega $ (rather than being the norm of a separator hyperplane) is a transformation from a $ n $-dimensional space to a one-dimensional space. With this interpretation, the intercept $ b $ is just a threshold that separates the transformed instances. LDA uses this interpretation and aims to find $ \omega:\mathbb{R}^n \to \mathbb{R} $ such that, in the transformed space, the distance between the centers of the classes is maximized while the spread of instances within the class is minimized. The early version of LDA \cite{fisher1936use} was proposed mainly for classification. In that version, it was assumed that the conditional probabilities $ P(\vec{x}^{(i)}|y^{(i)}=-1) $ and $ P(\vec {x}^{(i)}|y^{(i)}=1) $ ($y^{(i)}$ is the label of $\vec{x}^{(i)}$) are both normally distributed with mean and covariance parameters $ \left({\vec {\mu }}_{1},\Sigma_1\right) $ and $ \left({\vec {\mu }}_{2},\Sigma_2\right) $. Given these assumptions, Fisher \cite{fisher1936use} proved that $ \omega=(\Sigma_1+\Sigma_2)^{-1}(\vec{\mu}_2-\vec{\mu}_1) $ and $ k={\frac {1}{2}}{\vec {\mu }}_{2}^{T}\Sigma_2^{-1}{\vec {\mu }}_{2}-{\frac {1}{2}}{\vec {\mu }}_{1}^{T}\Sigma_1^{-1}{\vec {\mu }}_{1} $ leads to maximizing of $\frac{W^TS_BW}{W^TS_WW}$, where $S_W=(\Sigma_1+\Sigma_2)$ and $S_B=(\vec{\mu}_2-\vec{\mu}_1)(\vec{\mu}_2-\vec{\mu}_1)^T$. In this case, $ \omega $ is considered to be the norm of a hyperplane that discriminates the two classes and $ k $ shifts the hyperplane to be between the two classes, i.e., $ \vec{x}\omega^T>k $ if the instance $ \vec{x} $ belongs to class $ 1 $ (i.e., $\Omega_{\omega,k}$ seeks to maximize $\frac{W^TS_BW}{W^TS_WW}$, and $\mathcal{D}(\vec{x})=sign(\mathcal{M}_{\omega,k}(\vec{x}))=\vec{x}\omega^T-k$). 

MBLDA \cite{wang2016mblda} is a recent extension of LDA that aims to find a set of transformations so that the between-class scatter is maximized between each pair of the classes. The algorithm was shown to be effective for multiclass classification. For binary classification however, the algorithm is exactly the same as LDA.

\subsection{Evolutionary algorithms}
Evolutionary algorithms, EAs, (e.g., particle swarm optimization \cite{bonyadi2016review} and evolutionary strategy \cite{beyer2002evolution}) work based on a population of candidate solutions that are evolved according to some rules until they converge to an optimum solution. Each evolutionary algorithm has specific properties that confer advantages/disadvantages for specific applications. These methods aim to use information coded in each individual in the population (with the size $\lambda$) and update them to find better solutions. For example, evolutionary strategy (ES) generates new individuals using a normal distribution with the mean of the current location of the individual and an adaptive variance, calculated based on the distribution of "good" solutions. Covariance matrix adaptation evolutionary strategy (CMAES) employs a similar idea but updates the covariance matrix of the normal distribution (rather than the variance alone) to generate new instances, accelerating convergence to local optima. This idea takes into account non-separability of dimensions during optimization and hence is more successful when the variables are interdependent. Particle swarm optimization (PSO) follows a different approach where individuals (particles) combine their own experiences with others and calculate movement directions in the search space. See \cite{beyer2002evolution,bonyadi2016review} for detail of these methods.

\section{Proposed evolutionary-based classifier}
\label{sec:proposed}
In this section we describe $\Omega_\beta$ and $ \mathcal{D}$ for our proposed classifier (note that $ \mathcal{M}_\beta $ is linear), Optimal-margin Evolutionary Classifier (OEC). We first modify the definition of the 0-1 loss function in Eq. \ref{eq:0-1-loss} to overcome its sensitivity to imbalances in the number of instances in different classes. We then propose an algorithm that, in a one-dimensional space, finds a hyperplane minimizing the total loss using the modified 0-1 loss function while maximizing the margin between the hyperplane and the instances from each class (minimum empirical risk). Finally, we propose an evolutionary-based algorithm to extend the use of this algorithm to multi-dimensional classification. 

\subsection{Shortcomings of the 0-1 loss function} 
The total loss calculation using the 0-1 loss function (Eq. \ref{eq:0-1-loss}) is sensitive to imbalance in the number of instances, a significant issue in real-world classification problems \cite{branco2016survey}. The reason is that the number of misses (function $ h(.) $ in Eq. \ref{eq:0-1-loss}) are counted without taking into account the total number of instances in each class. Consider, for example, that there are two classes, the first class contains 200 instances and the second class contains only 10 instances. Also assume that hyperplane $ \mathcal{H} $ misclassifies all instances in the second class and classifies all instances in the first class correctly while hyperplane $ \mathcal{H}' $ misclassifies more than 10 instances from the first class and classifies all instances from the second class correctly. Using the 0-1 loss function and the total loss, hyperplane $ \mathcal{H}' $ is evaluated to be worse than $ \mathcal{H} $ because the number of misclassified instances is larger for $ \mathcal{H}' $, ignoring the fact that $ \mathcal{H} $ cannot classify the second class at all. In addition, two hyperplanes that provide the same total loss using the 0-1 loss function are evaluated to be the same, ignoring any empirical risk considerations. 

\subsection{Modified one-dimensional 0-1 loss function} \label{sec:alternativediscriminators}
To fix the sensitivity to imbalance in the 0-1 loss function, we minimize the summation of the ratio of misclassified instances rather than the \textit{number} of misclassified instances. As this ratio is in $ [0, 1] $ for all classes, it is independent of the number of instances, hence, this approach leads to removing the sensitivity to imbalance in the number of instances in the classes. We call this the 0-1 loss function with priors (LFP-0-1), defined by:
\begin{equation}
\label{eq:loss0-1-prior}
h(\vec{x}, y, <\omega, b>)=
\begin{cases}
\frac{1}{n_{1}} & y(x \omega^T +b) \le 0 \text{ and } y = 1\\
\frac{1}{n_{-1}} & y(x \omega^T +b) \le 0 \text{ and } y = -1\\
0 & y(x \omega +b)>0
\end{cases}
\end{equation}
The total loss value is calculated by Eq. \ref{eq:general-loss}, given $ \alpha $ and and the regularization function $ \mathcal{R}(.) $. 

Let us assume that the number of dimensions of the given instances ($ X $) is one, $ n=1 $. In this case, the optimal hyperplane that separates the classes (i.e., minimizes the total loss with LFP-0-1 loss function) is expressed by $ s \in \{-1,1\} $ and a threshold (intercept) $ b=t $. We define the \textit{optimal threshold}, optimal $ t $, as the threshold that minimizes the total loss with LFP-0-1 loss function. The value of $ s $ may negate the order of instances while the threshold $ t $ differentiates between them. We prove the following Remark, that will be used for our further discussions.

\begin{myRemark}
Let $ X $ be an $ m \times 1 $ matrix of one dimensional instances, and $ Y $ is the same size representing the class labels of the instances. Also, for a given threshold, $ t \in \mathbb{R} $, let the ratio of the instances in $ X $ from class $ -1 $ on the left hand side of $ t $ be $ l_{-1}(t) $ and the ratio of the instances from class 1 on left hand side of $ t $ be $ l_1(t) $. The optimal threshold has the minimum value of $ l_{-1}(t)+(1-l_1(t)) $ or $ l_1(t)+(1-l_{-1}(t)) $, over all possible possible $ t $. 
\end{myRemark}
\begin{proof}
	There are two cases: the instances on the left hand side of $ t $ are classified as class $ -1 $ or $ 1 $. In the first case, the instances to the left hand side of $ t $ (inclusive) are classified to class $ -1 $, $ 1-l_{-1}(t) $ and $ l_1(t) $ are the ratio of misclassification of the instances from class $ -1 $ and $ 1 $, respectively, at the threshold $ t $. Hence, $ (1-l_{-1}(t))+l_1(t) $ is the total ratio of misclassification of instances belong to the class $ -1 $ plus the ratio of the misclassification of instances in the class $ 1 $ at the threshold $ t $. In the second case, the instances on the left hand side of $ t $ are classified to class $ 1 $, $ l_{-1}(t)+(1-l_1(t)) $ is the ratio of misclassification of instances in the class $ 1 $ plus the ratio of the misclassification of instances in the class $ 1 $ at the threshold $ t $. Thus, if either $ l_{-1}(t)+(1-l_1(t)) $ or $ (1-l_{-1}(t))+l_1(t) $ is minimum at a given $ t $ then that $ t $ is an optimal threshold.
\end{proof}

\begin{figure}
	\centering
	\includegraphics[width=0.49\textwidth]{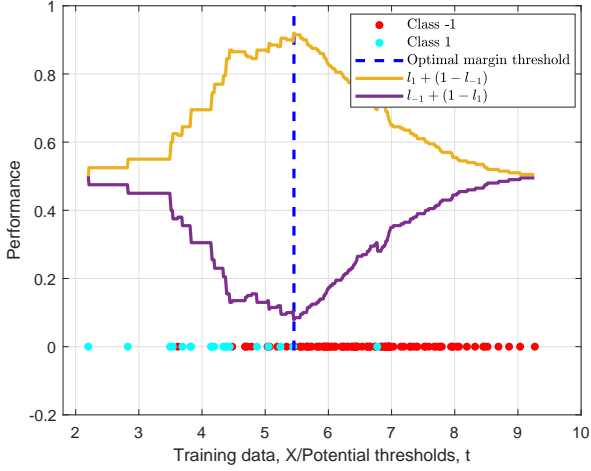} 
	\caption{Value of $ l_{-1}(t)+(1-l_1(t)) $ and $ l_1(t)+(1-l_{-1}(t)) $ (vertical axis) vs the threshold $ t $ (horizontal axis). The horizontal axis also shows the coordinate of the training data, $ X $, in a one-dimensional space. Note that the value "0" for the vertical coordinate of the instances is arbitrary and it is for demonstration purposes only.}
	\label{fig:optimal_threshold}
\end{figure}

Figure \ref{fig:optimal_threshold} shows the value of $ l_{-1}(t)+(1-l_1(t)) $ and $ l_1(t)+(1-l_{-1}(t)) $ for different $ t $. 

The optimal threshold that optimizes the total loss with LFP-0-1 loss function might not be unique. This can be observed in Fig. \ref{fig:optimal_threshold}, where $ l_1(t)+(1-l_{-1}(t)) $ indicates many plateaus for different values of $ t $. The reason is that any value for $ t $ between two consecutive instances (recall that $ X $ is one-dimensional) would lead to a constant $ l_1(t) $ and $ l_{-1}(t) $, hence, the value of $ l_{-1}(t)+(1-l_1(t)) $ and $ l_1(t)+(1-l_{-1}(t)) $ are constant between each instance and the next (i.e., $ l_{-1}(x_i)+(1-l_1(x_i))=l_{-1}(t)+(1-l_1(t)) $ for all $ t \in [x_i, x_{i+1}) $). Therefore: 
\begin{enumerate}
	\item To find an optimal $ t $, we only need to test the value of $ l_{-1}(t)+(1-l_1(t)) $ and $ l_1(t)+(1-l_{-1}(t)) $ at the location of instances, i.e., for $ t=x_i $ for all $ i $, where $ x_i $ is the $ i^{th} $ instance in $ X $. 
	\item The optimal threshold with minimum risk in the one-dimensional space is half-way between the instance that provides the minimum total loss ($ l_{-1}(t)+(1-l_1(t)) $ or $ l_1(t)+(1-l_{-1}(t)) $) and the next instance. The reason is that this threshold is the furthest from the closest instances from each class. This is called the \textit{optimal-margin threshold}, $ t' $, throughout the paper, and is unique. 
\end{enumerate}

We now propose Algorithm \ref{alg:OPTdiscrim} to obtain the optimal-margin threshold, $ t $, and $ s \in \{-1,1\} $ to distinguish between two classes in a one dimensional space. The main idea is to iterate over all instances ($ x_i\in X $, where $ X $ is sorted), sorted in the one dimensional space, and calculate $ l_{-1}(x_i) $ and $ l_1(x_i) $. Ultimately, the goal is to find the "best instance" that, if used as a threshold ($ t $), the total misclassification percentage is minimized, that is either $ l_{-1}(t)+(1-l_1(t)) $ or $ l_1(t)+(1-l_{-1}(t)) $ depending on the order of classes. To maximize the margin, $ r $, we set $ t' $ to half way between the best instance and the next instance. The time complexity of this algorithm is in $ O(m\log(m)) $ (to sort the instances), where $ m $ is the number of instances. 

\begin{algorithm}[tb]
	\caption{Optimum margin threshold
		\textbf{Input}: $ X $: the training set, $ m \times 1 $; $ Y $: the class labels, $ m \times 1 $
		\textbf{Output}: $ t' $: the optimal margin threshold; $ s $: a coefficient; $ p $: the performance; $ r $: size of margin}
	\label{alg:OPTdiscrim} 
	\begin{algorithmic}[1]
		\STATE $ X $=sort($ X $), and update $ Y $ accordingly
		\STATE $ l_{-1}(x_0)=0 $, $ l_1(x_0)=0 $, $ p=0 $, $ t'=-\inf $
		\STATE $ k_{-1} $ = number of instances in class -1	
		\STATE $ k_1 $ = number of instances in class 1	
		\FOR {i=1 to number of instances $ - 1 $}
			\IF {$ x_i $ (the i$ ^{th} $ instance in $ X $) is in class $ -1 $}
				\STATE $ l_{-1}(x_i)=l_{-1}(x_{i-1})+\frac{1}{k_{-1}} $
			\ELSE 
				\STATE $ l_1(x_i)=l_1(x_{i-1})+\frac{1}{k_1} $
			\ENDIF
			\STATE $ a_{-1}=l_{-1}(x_i)+(1-l_1(x_i)) $, $ a_{1}=l_1(x_i)+(1-l_{-1}(x_i)) $
			\IF {$ a_{-1}>p $}
				\STATE $ t=\frac{x_i+x_{i+1}}{2} $, $ s=1 $, $ p=a_{-1} $
				\STATE $ r= \frac{|x_i-x_{i+1}|}{2} $
			\ENDIF
			\IF {$ a_1>p $}
				\STATE $ t=-\frac{x_i+x_{i+1}}{2} $, $ s=-1 $, $ p=a_1 $
				\STATE $ r= \frac{|x_i-x_{i+1}|}{2} $
			\ENDIF
		\ENDFOR		
	\end{algorithmic}
\end{algorithm}

This algorithm is able to find $ t' $ and $ s $ in a one dimensional space that optimizes Eq. \ref{eq:0-1-loss} with $ h(.) $ defined by Eq. \ref{eq:loss0-1-prior} \footnote{The proof is trivial and can be achieved by contradiction.}. The purpose of $ s $ is to make sure that the instances from class +1 are always at the right hand side of the threshold, $ t $, and ultimately $ t' $. This is arbitrary and does not have any impact on the final results, however, it standardizes the expectation of where the instances of different classes would place after the optimization. The value of $ p $ indicates the average performance in terms of the ratio of correctly classified instances. A new instance $ x $ is classified to class $ -1 $ if $ s x \le t' $, and to class $ +1 $ otherwise. 

\subsection{Evolutionary classifier: binary classification}
\label{sec:evo-binary}
By definition, we seek  $ \omega: \mathbb{R}^n \to \mathbb{R} $ so that $ X\omega^T $ is separable by $ \mathcal{D}(.) $. Geometrically, $ \omega $ is a line that passes through the center of the coordinates system (intercept equal to zero). As $ X\omega^T $ represents the projected instances on the line $ \omega $, we seek the best $ \omega $ such that the projection of instances on it ($ X\omega^T $, that is one-dimensional) maximizes the separability, measured by $ \mathcal{D}(.) $. Because $ X\omega^T $ is one-dimensional, we use Algorithm \ref{alg:OPTdiscrim} as our $ \mathcal{D}(.) $ to discriminate between classes. Hence, the aim is to find an $ \omega $ for which the performance and margin calculated by Algorithm \ref{alg:OPTdiscrim} using $ X\omega^T $ and $ Y $ (the classes) is maximized. For this objective, finding the best $ \omega $ is not a convex optimization problem, hence, an EA is a good choice for optimization purposes. The objective value for EA is calculated by Algorithm \ref{alg:objective}.

\begin{algorithm}[tb]
	\caption{Objective function
		\textbf{Input}: $ X $: the $ m \times n $ training data; $ Y $: the class labels, $ m \times 1 $; $ \beta $: the transformation; $ \alpha $: $ L_1 $ regularization factor
		\textbf{Output}: $ z $: objective value}
	\label{alg:objective} 
	\begin{algorithmic}[1]
		\STATE $ \hat{X}=X\omega^T $
		\STATE $ [t',s,p,r]= $Algorithm \ref{alg:OPTdiscrim} ($ \hat{X} $, $ Y $)
		\IF {$ p<1 $}
			\STATE $ z=p $
		\ELSE
			\STATE $ z=1+r $
		\ENDIF
		\STATE $ z=z+\alpha||\beta||_1 $
	\end{algorithmic}
\end{algorithm}

If the transformed instances are separable, then the performance $ p $ increases to $ 1 $. In this case, the final performance is set to $ 1+r $ to ensure that, among all possible $ \omega $ that separate the instances, the $ \omega $ which maximizes the margin is selected. The maximum margin provides a lower risk on classification of unseen instances. 

In the cases where the instances are linearly separable, the margin $ r $ may increase only because $ ||\omega||_2 $ increases. The reason is that, by increasing $ ||\omega||_2 $, the distance between all instances increases, leading to an "illusion" of a larger margin, $ r $. Hence, in all of our EA algorithms, we normalize the candidate solutions at each iteration to ensure that the solutions do not grow unbounded, that is equivalent to $ L_2 $ regularization. We call this algorithm the Optimal-margin Evolutionary Classifier (OEC). The $ L_1 $ regularization has been considered in our calculations by optimizing $ z+\alpha||\omega||_1 $ rather than $ z $ itself. 

After optimization of $ \omega $, an instance $ \vec{x} $ is classified to class $ -1 $ if $ s \vec{x}\omega^T < t' $, and to class $ 1 $ otherwise. Thus, we may assume that $ s\vec{x}\omega^T-t' $ is the equation for the separator hyperplane found by our algorithm, where $ s\omega $ is the normal vector of the hyper plane and $ -t' $ is the intercept \footnote{OEC source codes in Java, Matlab, and Python are available online at \url{https://github.com/rezabonyadi/LinearOEC}}.

\subsection{Evolutionary classifier: multi-class classification}
We use the one-vs-one method \cite{hsu2002comparison} to extend OEC to work with multi-class classification. Here, we classify the instances in each pair of classes as a binary classification problem. For each class pairs $ i $ and $ j $, OEC finds a transformation $ \omega_{i,j} $, intercept $ t'_{i,j} $, and coefficient $ s_{i,j} $, that best separates those classes (see Algorithm \ref{alg:one-vs-one}).  Algorithm \ref{alg:one-vs-one} provides exactly $ \frac{c(c-1)}{2} $ transformations, each optimally (using the objective Algorithm \ref{alg:objective}) calculated to separate one pair of classes. 

\begin{algorithm}[tb]
	\caption{Multiclass OEC
		\textbf{Input}: $ X $: the $ m \times n $ training data, $ Y $: class labels
		\textbf{Output}: $ \omega $: the set of best transformations; $ t' $: the set of best thresholds; $ e $: set of transformation coefficients; $ p $: the average estimated performance}
	\label{alg:one-vs-one} 
	\begin{algorithmic}[1]
		\FOR {i=1 to number of classes - 1}
			\FOR {j=i+1 to number of classes}
				\STATE $ X_{i,j} = $ all instances in $ X $ that belong to class $ i $ or $ j $
				\STATE $ Y_{i,j}= $ all class labels of instances in classes $ i $ or $ j $, translated to labels $ -1 $ and $ 1 $
				\STATE Use an evolutionary algorithm to find the best $ \omega_{i,j} $, a $ n \times 1 $ transformation, with objective in Algorithm \ref{alg:objective}($ X_{i,j} $, $ Y_{i,j} $, $ \omega_{i,j} $)
				\STATE Calculate $ \hat{X} = X\omega_{i,j}^T $,
				\STATE Find optimum $[t'_{i,j},s_{i,j},p_{i,j}, r_{i,j}] =$ Algorithm \ref{alg:OPTdiscrim}($ \hat{X} $, $ Y_{i,j} $),  
			\ENDFOR	
		\ENDFOR	
	\end{algorithmic}
\end{algorithm}

To classify a new instance, we simply use a voting system in which the instance is transformed by all transformations (all $ \omega_{i,j} $) and the class is calculated based on the votes received by each $ t'_{i,j} $ and $ s_{i,j} $ (see Algorithm \ref{alg:one-vs-one-preict}). 

\begin{algorithm}[tb]
	\caption{Get Class
		\textbf{Input}: $ x $: the $ 1 \times n $ training data; $ \omega $: A $ c \times c $ matrix of transformations, $ n \times 1 $ each; $ t' $: A $ c \times c $ matrix of thresholds; $ s $: A $ c \times c $ matrix, each value in $ \{-1, 1\} $
		\textbf{Output}: the class label $ e $}
	\label{alg:one-vs-one-preict} 
	\begin{algorithmic}[1]
		\STATE $ E=<0,...,0> $ with the length $ c $.
		\FOR {i=1 to number of classes - 1}
			\FOR {j=i+1 to number of classes}
				\STATE if $ s_{i,j}x \omega_{i,j}^T<t' $ then $ y=-1 $, else $ y=1 $.
				\STATE if $ y=1 $ then $ E_i=E_i+1 $ else $ E_j=E_j+1 $.
			\ENDFOR	
		\ENDFOR	
		\STATE $ e=\argmax \limits_i \{E_i\} $.
	\end{algorithmic}
\end{algorithm}

\subsection{A visual overview through an example}
\label{sec:cisualoverview}
\begin{figure}
	\centering
	\includegraphics[width=0.49\textwidth]{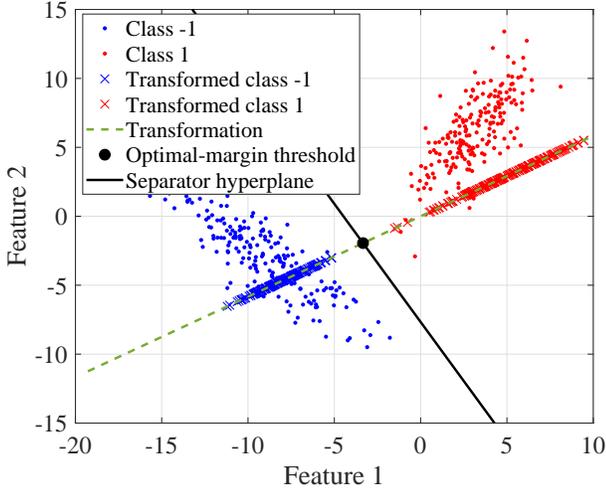} 
	\caption{After the transformation, the instances are mapped into a line, represented by $ \omega $. Alternatively, they are separable by the separator hyperplane.}
	\label{fig:OEC-example-real}
\end{figure}

We generated a synthetic dataset for a binary classification problem for the purpose of demonstrating how OEC works. We used a multivariate normal distribution,\footnote{The covariance matrices and the means have been selected by some trial to illustrate the procedure of OEC as clear as possible and they do not have any other specific characteristics.} with covariance matrix of $ \begin{bmatrix}
0.87&-0.5\\
1.5&2.6\\
\end{bmatrix}$ and $ \mu=<3,6> $ for class $ 1 $, and with covariance matrix 
$ \begin{bmatrix}
2.83&-2.83\\
0.71&0.71\\
\end{bmatrix} $ and $ \mu=<-9,-3> $ for class $ -1 $, and characteristics are shown in Figure \ref{fig:OEC-example-real}. 

We then found the optimal $ \omega $ by an EA (CMAES in this example), as well as $ s $ and $ t' $ by Algorithm \ref{alg:OPTdiscrim}, described in Section \ref{sec:evo-binary}. Note, $ \omega $ found by this method transforms instances from an $ n $-dimensional space to a one-dimensional space where the instances are separable by a discriminator threshold, $ t' $. An alternative view is that the instances are separable by a hyperplane, $ s\vec{x}\omega^T-t' $. Figure \ref{fig:OEC-example-real} shows the transformation and the separator hyperplane.

\begin{figure}
	\centering
	\includegraphics[width=0.49\textwidth]{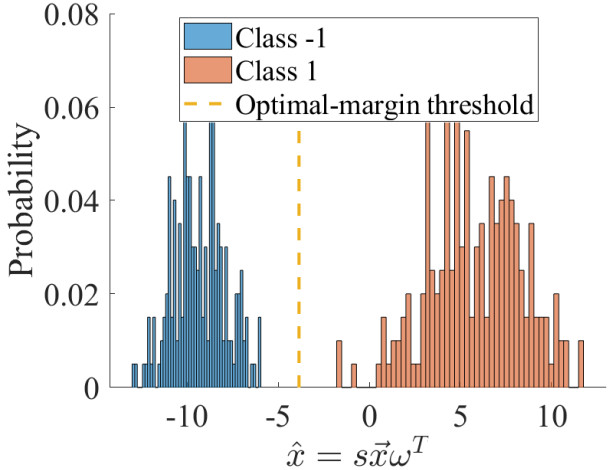}     
	\centering    
	\caption{The distribution of the instances after optimization of $ \omega $. The yellow dashed line indicates the optimal-margin threshold found by Algorithm \ref{alg:OPTdiscrim}.}
	\label{fig:OEC-example-hist}
\end{figure}

The histogram of the transformed instances after optimizing $ \omega $ is shown in Fig. \ref{fig:OEC-example-hist}. The optimal margin threshold has been also shown in the figure. 

It is clear that the proposed OEC is able to find a high-quality separator hyperplane to distinguish between classes.

\section{Experiments and results}
\label{sec:experiment}
We compare OEC with six other classification methods over 19 standard classification problems in this section.

\subsection{Outline of comparisons}
Here we introduce the datasets, pre-processes, and algorithm specific settings used in the comparisons. 

\subsubsection{Algorithm settings}
We compared the performance of OEC with SLA, SVM, SMP, DEL, WSV, and MBA. We used MATLAB 2017a for implementations and tests.\footnote{The source code for OEC is available as a supplementary to this article.} The parameter values for these methods were as follows: 
\begin{itemize}
	\item DEL: 400 iterations of optimization, $ \lambda=.5 $ (regularization parameter), $ \mu=1e-8 $ (tolerance). 
	\item WSV: $c_1=0.1$, $c_2=0.1$, $c_3=0.1$, $c_4=0.1$ (the coefficients), as suggested in the source code. 
	\item SLA and SMP: none (no parameters).
	\item SVM: $ \lambda=1 $ (the margin/accuracy balance factor). 
	\item MBA: $ \lambda = 0.5 $ (the regularization factor).
	\item OEC: number of individuals $ \lambda=4+3\log(n) $, number of offspring (in ES and CMAES) was $ \lambda/2 $, maximum number of iterations $ 150\log(n+1) $.
\end{itemize}

\subsubsection{Datasets and performance measures}
We used 12 standard binary classification datasets (namely, Breast cancer (BC) , Crab gender (CG), Glass chemical (GC), Parkinson (PR), Ionosphere (IS), Pima Indians diabetes (PF), German credit card (GR), Vote (VT), Madelon (MD), Hill-Valley without noise (HV), Hill-valley with noise (HN), Seizure recognition (SR)) and seven multi-class classification datasets (Iris (IR), Italian wine (IW), Thyroid function (TF), Yeast dataset (YD), Red wine quality (RW), White wine quality (WW), and Handwritten Dataset (HD))\footnote{All of these datasets are available online at https://archive.ics.uci.edu/ml/datasets.html} to compare methods. The main characteristics of these datasets have been provided in Table \ref{Tab:dbs1}. These datasets are used frequently as standard benchmarks in machine learning studies. 
	
\begin{table}[]
	\centering
	\caption{The datasets used for comparison purposes in this paper. $ n $ is the number of variables and $ c $ is the number of classes in each dataset. The number of instances in each class has been reported in the last column.}
	\label{Tab:dbs1}
	\begin{tabular}{|l|l|l|l|}
		\hline
		\begin{tabular}[c]{@{}l@{}}\textbf{Dataset}\\ \textbf{name}\end{tabular}
		& \textbf{$ n $} & \textbf{$ c $} & \begin{tabular}[c]{@{}l@{}}\textbf{Number of instances} \textbf{in each class}\end{tabular}\\ \hline
		BC & 9 & 2 & $ <458,241> $ \\ \hline
		CG & 6 & 2 & $ <100,100> $ \\ \hline
		GC & 9 & 2 & $ <51,163> $ \\ \hline
		PR & 22 & 2 & $ <48,147> $ \\ \hline
		IS & 32 & 2 & $ <225,126> $ \\ \hline
		PD & 8 & 2 & $ <268,500> $ \\ \hline
		GR & 24 & 2 & $ <300,700> $ \\ \hline	
		VT & 8 & 2 & $ <192,242> $ \\ \hline		
		MD & 500 & 2 & $ <1300,1300> $ \\ \hline			
		HV & 100 & 2 & $ <606,606> $ \\ \hline
		HN & 100 & 2 & $ <606,606> $ \\ \hline
		SR & 178 & 2 & $ <9200,2300> $ \\ \hline	
		IR & 4 & 3 & $ <50,50,50> $ \\ \hline
		IW & 13 & 3 & $ <59,71,48> $ \\ \hline	
		TF & 21 & 3 & $ <166,368,6666> $ \\ \hline
		YD & 8 & 10 & $ <463,5,35,44,51,163,244,429,20,30> $ \\ \hline
		RQ & 11 & 6 & $ <10,53,681$ $,638,199,18> $ \\ \hline
		WQ & 11 & 7 & $ <20,163,1457,2198,880,175,5> $ \\ \hline
        HD & 784 & 10 & \begin{tabular}[c]{c} $ <6903,7877,6990,7141,6824,6313 ,$ \\ $6876,7293,6825,6958> $ \end{tabular} \\ \hline
	\end{tabular}	
\end{table}

We structured our comparisons as follows:
\begin{enumerate}
	\item We first compared three evolutionary algorithms, namely PSO, ES, and CMAES, for optimization of the objective function in Algorithm \ref{alg:objective}. This comparison provided insight on the best choice of optimization algorithm for OEC.
	\item We compared OEC with SLA and optimized SVM on binary classification problems. The aim of this comparison was to show how close OEC results are to near-optimal solutions SLA and optimized SVM find. We run the algorithms for 100 times within each we selected 70\% of instances from each class for training and the rest for testing (stratified sampling\cite{kohavi1995study}). The training and testing sets remained constant for all methods. 
	\item We tested the sensitivity to the imbalanced number of instances in classes of OEC with SLA and optimized SVM on a synthetic dataset.
	\item We compared OEC with state-of-the-art classification algorithms, namely SVM, SMP, WSV, DEL, and MBA, on binary classification problems with and without noise and outliers. The aim of this comparison was to show whether OEC provides high-quality generalizable solutions to standard binary optimization problems within a practical amount of time.
	\item We tested the impact of $ L_1 $ regularization on the found weights ($ \omega $). 
	\item We compared OEC against SVM, MBA, DEL, and WSV on multi-class classification problems. We did not include SMP and SLA as they require a long time to achieve results on these datasets.
\end{enumerate}

Area Under the Curve (AUC) of the Receiver Operating Characteristic (ROC) \cite{hanley1982meaning} curve were used as a performance measure. The training dataset was normalized to ensure each feature has a zero mean and a unit variance. The transformation computed for normalization was then applied to the test set to ensure consistency between train and test domain. 

\subsection{Comparison between different EAs}
\begin{figure}
	\centering
	\includegraphics[width=0.49\textwidth]{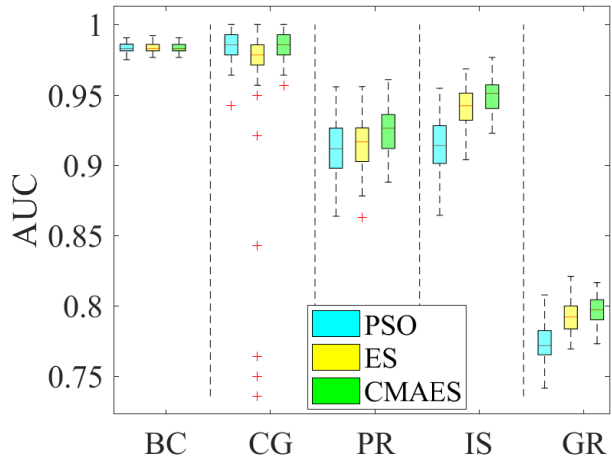} 
	\caption{Comparison between PSO, ES, and CMAES for binary classification. Results are the performance of each method for training.}
	\label{fig:EA_compare}
\end{figure}

We tested the performance of three EAs, namely PSO, ES, and CMAES, on five binary classification problems. Figure \ref{fig:EA_compare} shows that CMAES outperforms ES and PSO in terms of AUC. In terms of running time, however, PSO and ES were the fastest (with no significant difference, $ p > 0.3 $) while CMAES was the slowest, as expected. 

We used CMAES in the rest of this paper for optimization purposes.

\subsection{Comparison with near-optimal discriminative classifiers}
We compared OEC with SLA, the fastest algorithm proposed in \cite{nguyen2013algorithms} to find a near-optimal separator hyperplane, and the optimized SVM in Table \ref{Tbl:comparisonResultsOptimal}. Optimized SVM was the standard SVM for which the parameters were optimized using the Bayesian optimization \cite{Gelbart:2014:Bayes}. OEC significantly outperformed SLA and optimized SVM in all tested datasets in the training set. For the testing sets, OEC is significantly better than SLA in two datasets and significantly better than optimized SVM in three datasets. This indicates that, given the training dataset, OEC finds a better hyperplane to distinguish between classes in comparison to SLA and optimized SVM. Also, the optimal-margin indeed improved the generalization ability of the algorithm to classify unseen cases correctly, evidenced by the better performance of OEC in the testing dataset. In terms of the running time, OEC outperformed both SLA and optimized SVM in all cases by taking less than 0.5 seconds for classification in all tested cases. 

\begin{table}[]
	\centering
	\caption{Comparison with SLA and optimized SVM over 100 runs. Times are in seconds. The character "*" on a value indicates that the OEC corresponding result was significantly better than this reported value (t-test, $ p < 0.05 $). The character "-" indicates that this result was significantly better than that of OEC's. The character "-" indicates that the result of OEC was statistically the same as this reported value.}
	\label{Tbl:comparisonResultsOptimal}
	\begin{tabular}{|l|l|l|l|l|l|l|}
		\hline
		\textbf{Algorithm} & \textbf{BC} & \textbf{CG} & \textbf{GC} & \textbf{PR} & \textbf{IS} & \textbf{PD} \\ \hline
		SLA-Time & 12.5* & 1.67* & 3.5* & 5.1* & 6.4* & 35.7*  \\\hline
		SLA-Train & 98.09* & 96.55* & 93.7* & 84.14* & 93.61* & 77.05*  \\\hline
		SLA-Test & 96.66- & 95- & 86.68* & 77.27* & 83.53+ & 71.62-  \\ \hline
		
		SVM-Time & 52.9* & 16.1* & 19.9* & 20.6* & 26.7* & 44.5*  \\ \hline
		SVM-Train & 96.81* & 96.44* & 92.43* & 82.92* & 93.2* & 73.9*  \\ \hline
		SVM-Test & 96.11* & \textbf{95.31}- & 87.21* & 77.74* & \textbf{84.32}+ & 72.30-  \\ \hline
		
		OEC-Train & \textbf{0.2} & \textbf{0.1} & \textbf{0.1} & \textbf{0.3} & \textbf{0.4} & \textbf{0.2}  \\ \hline
		OEC-Train & \textbf{98.42} & \textbf{98.44} & \textbf{97.12} & \textbf{92.41} & \textbf{95.2} & \textbf{79.88}  \\ \hline
		OEC-Test & \textbf{96.71} & 95.12 & \textbf{90.22} & \textbf{79.72} & 82.32 & \textbf{72.74}  \\ \hline
	\end{tabular}
\end{table}

\subsection{Class-imbalance test}
\begin{figure}
	
	\begin{tabular}{cc}
		\centering
		\includegraphics[width=0.23\textwidth]{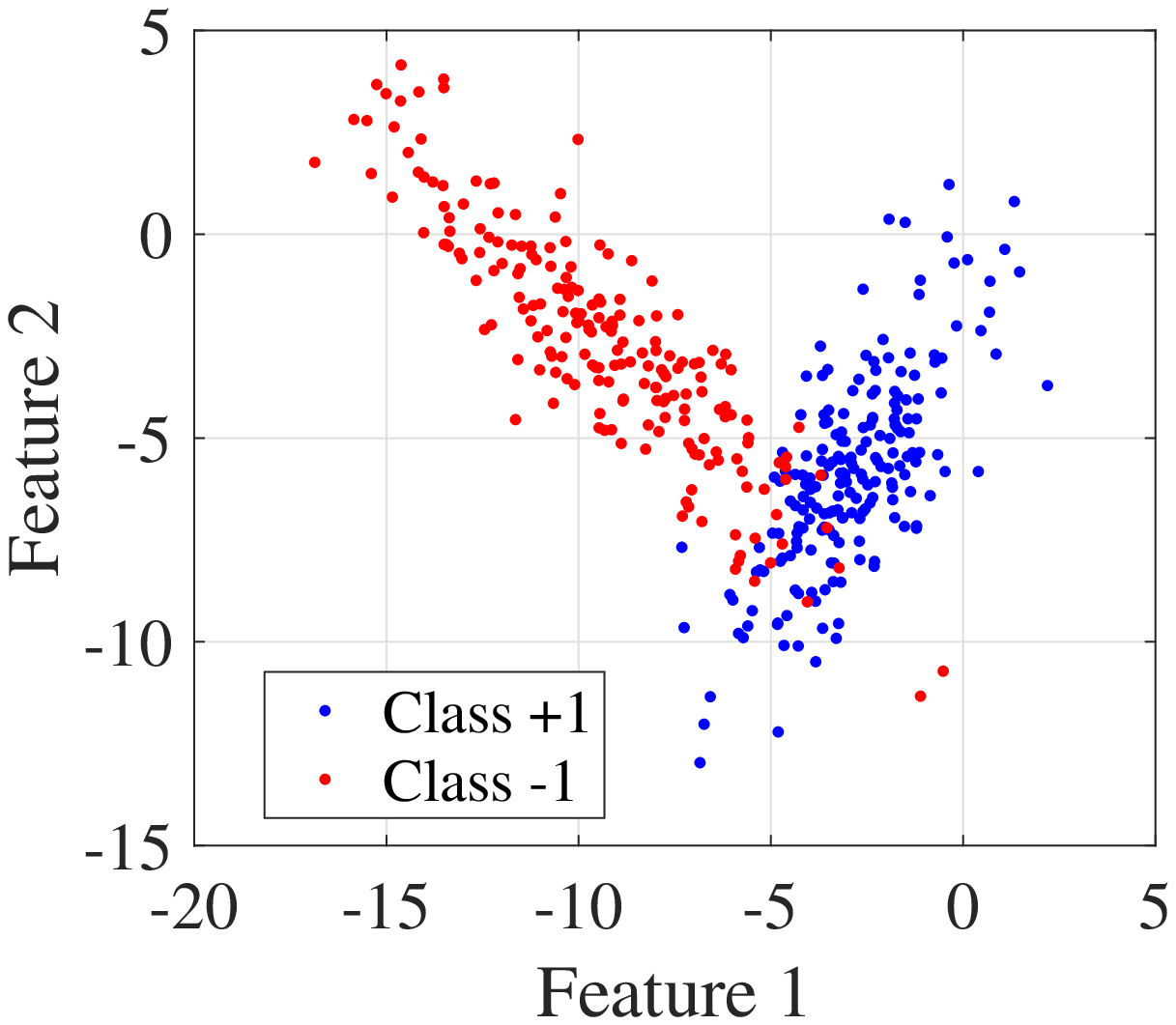} & \includegraphics[width=0.23\textwidth]{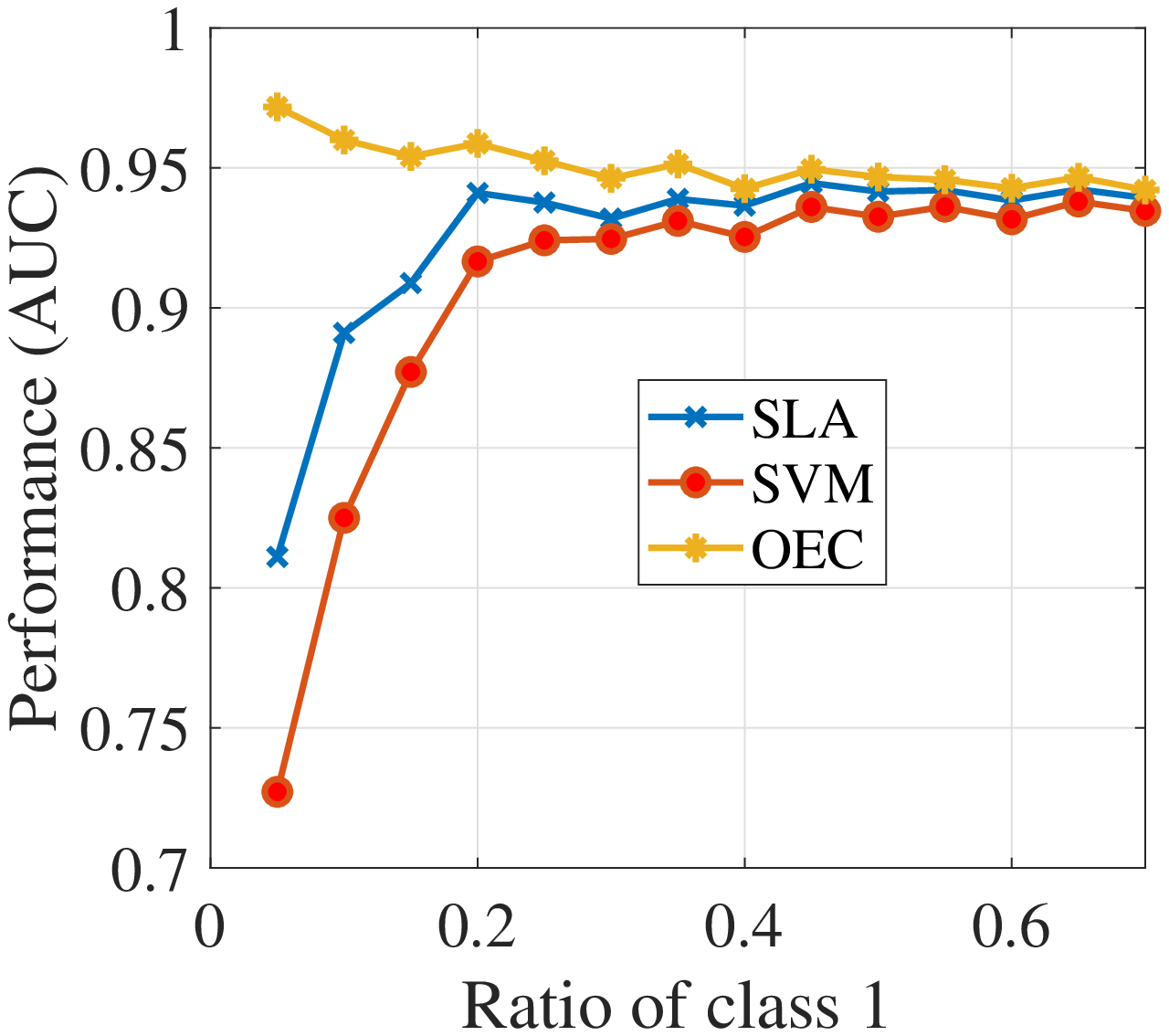} \\
		(a) & (b)		
	\end{tabular}
	\caption{Sensitivity to imbalance number of instances in the classes. (a) is the tested dataset and (b) is the performance of different algorithms when the ratio of instances from class $ 1 $ changed for training.}
	\label{fig:class_imbalance}
\end{figure}

We tested the efficiency of OEC in classifying a dataset with imbalanced numbers of instances in each class. We generated a synthetic binary classification dataset using the procedure described in Section \ref{sec:cisualoverview}, where we changed the mean ($ \mu $) of class $ +1 $ to $ \mu=<-3,-6> $. With this setting, the instances in the class $ +1 $ and $ -1 $ would overlap (see Fig \ref{fig:class_imbalance}(a)). We then generated the training set using 70\% of instances from class $ -1 $ while changed the ratio of instances from class $ +1 $ appeared in the traning set in the range of 5\% to 70\% with the step size of 5\%. Results in Fig \ref{fig:class_imbalance}(b) show that both SLA and SVM are not effective when the number of instances in the classes is imbalanced. This, however, is not true for OEC where the performance was consistent for different ratios of instances in class $ -1 $.

\subsection{Binary classification with and without noise and outlier}
We also compared OEC with state-of-the-art classification algorithms, namely DEL, WSV, SMP, SVM, and MBA, in Table \ref{Tbl:comparisonResultsDisc-summary} on both noisy and non-noisy data (complete report can be found in Appendix). For non-noisy cases (row "0\% noise"), each column of the row Train in the table indicates the number of datasets for which OEC performed significantly better (or worse, in parentheses) than the method in that column for the training set. The same results for the test set have been reported in the row Test. Results indicate that compared to all other algorithms, OEC finds a significantly better hyperplane to separate classes, given the training sets, in all 12 datasets except for one dataset in comparison to WSV, where the performance of OEC and WSV was similar ($ p>0.05 $)\footnote{The dataset was the SR dataset; p=0.07 could be interpreted as being close to statistical significance.}. OEC was also significantly better than other methods in the testing sets in most cases. 

We used the procedure in \cite{nguyen2013algorithms} to introduce noise and outliers to the binary classification datasets in Table \ref{Tab:dbs1}. The noise was generated as a random value in $ [min(x_i)-0.5(max(x_i)-min(x_i)), max(x_i)+0.5(max(x_i)-min(x_i))] $ (uniform distribution), where $ min(x_i) $ is the minimum value for the $ i^{th}  $ variable in the training set and $ max(x_i) $ is the maximum value for the $ i^{th}  $ variable in the training set. This formulation ensures that the instances are noisy while outliers are also likely to be generated. We selected 15\% of instances from class $ +1 $ randomly and perturbed them using this procedure. Results in Table \ref{Tbl:comparisonResultsDisc-summary}, row noise "15\%", indicate that the number of datasets for which OEC outperforms other classifiers has increased. This indicates that the performance drop in other methods was more severe than that of OEC.

\begin{table}[]
	\centering
	\caption{The values indicate the number of datasets (over 12 in total) for which OEC outperforms other methods significantly ($ p<0.05 $) in the training and testing sets. The values in parentheses show the number of datasets for which OEC was outperformed by another method significantly. Results are with noise (15\%) and without noise.}
	\label{Tbl:comparisonResultsDisc-summary}
	\begin{tabular}{|l|l|l|l|l|l|l|}
		\hline
		Noise & Set & \textbf{DEL} & \textbf{WSV} & \textbf{SMP} & \textbf{SVM} & \textbf{MBA} \\ \hline
		\multirow{2}{*}{0\%} & Train & 12(0) & 11(0) & 12(0) & 12(0) & 12(0) \\\cline{2-7} 
		& Test & 7(1) & 6(4) & 6(2) & 9(1) & 7(0) \\ \hline
		\multirow{2}{*}{15\%} & Train & 12(0) & 12(0) & 12(0) & 11(0) & 12(0)  \\ \cline{2-7} 
		&Test & 10(1) & 10(1) & 8(3) & 9(1) & 10(1)  \\ \hline
		\end{tabular}
\end{table}

\begin{figure}
	\centering
	\includegraphics[width=0.49\textwidth]{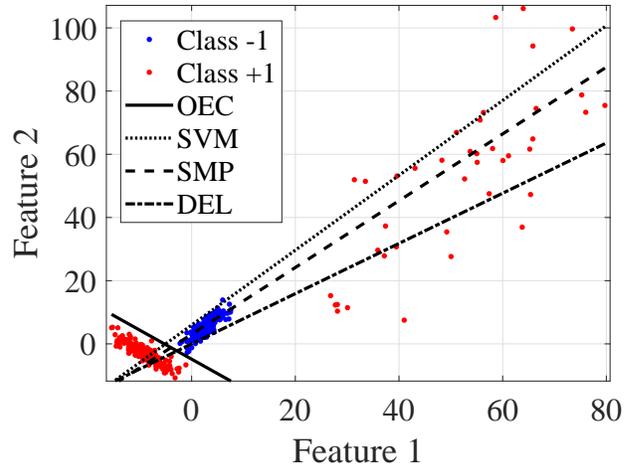} 
	\caption{A synthetic two-dimensional binary classification problem with outlier. All methods except OEC found poor quality separator hyperplanes except.}
	\label{fig:outliers}
\end{figure}

To visualize the impact of outlier on OEC, DEL, SMP, and SVM, we generated a synthetic dataset using the procedure described in Section \ref{sec:cisualoverview} and then added outlier instances to the class $ +1 $. The outliers were generated randomly using a normal distribution with covariance matrix $ \begin{bmatrix}
8.66&-5\\
15&25.98\\
\end{bmatrix}$ and $ \mu=<50,50> $. We compared the results of OEC for this dataset against SVM, SMP, and DEL in Fig. \ref{fig:outliers}. 
The hyperplane (line in 2-dimensional space) found by DEL, SMP, and SVM have been influenced significantly by outliers. OEC, however, still maintained its good performance and found a relatively better separator line. The AUC of the training set after optimization was 92.6, 66.5, 43.6, and 57.2 for OEC, SMP, SVM, and DEL, respectively. 
 
\subsection{Impact of $ L_1 $ regularization}
\begin{figure}
	\centering
	\includegraphics[width=0.49\textwidth]{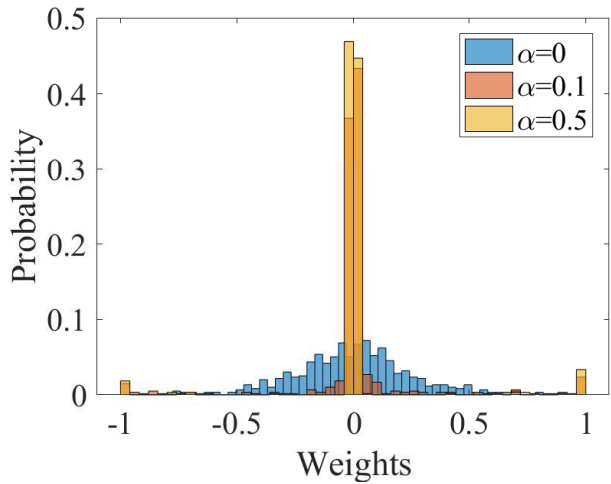} 
	\caption{Distribution of the weights with $ L_1 $ regularization factor ($ \alpha $) equal to 0, 0.1, and 0.5. Weights are more distributed around 0, 1, and $ -1 $ when the regularization factor is larger than zero.}
	\label{fig:l1-regularization}
\end{figure}

Incorporating the $ L_1 $ regularization term to the optimization objective leads to a preference over weights that are closer to 0, 1, or $ -1 $ \cite{bishop2006pattern}. This makes the $ L_1 $ regularized solutions a good candidate for feature selection. Hence, we set the regularization factor $ \alpha $ to test its impact on the found solutions. We ran OEC with $ \alpha \in \{0, 0.1, 0.5\}$ and applied the algorithm to the binary classification problems in Table \ref{Tab:dbs1}. We ran the algorithm 10 times on each dataset and recorded the found $ \omega $. Figure \ref{fig:l1-regularization} shows the histogram of weights after optimization.

It is clear that the weights found under larger $ \alpha $ are more distributed around 0, 1, and $ -1 $. Note, however, that the performance of the algorithm dropped by 2\% for $ \alpha=0.1 $ and 5\% for $ \alpha=0.5 $ in average.

\subsection{Comparison with state-of-the-art multi-class classification methods}
We used the one-vs-one method to extend binary classification algorithms to deal with multiclass problems. Results have been summarized in Table \ref{Tbl:comparisonResultsDiscMulticlass} (see Appendix for details of these results). These results show that the number of datasets for which OEC performs significantly better than other methods, in training and testing datasets, is larger than the number of datasets for which OEC performs significantly worse. This indicates that OEC outperforms other tested methods in solving multiclass classification problems using the one-vs-one method.

\begin{table}[]
	\centering
	\caption{One-vs-one strategy on 7 datasets.}
	\label{Tbl:comparisonResultsDiscMulticlass}
	\begin{tabular}{|l|l|l|l|l|}
		\hline
		Dataset & \textbf{DEL} & \textbf{WSV} & \textbf{MBA} & \textbf{SVM} \\ \hline
		\textbf{Train} & 7(0) & 7(0) & 5(0) & 5(0)\\ \hline
		\textbf{Test} & 7(0) & 5(2) & 5(2) & 3(2)\\ \hline
	\end{tabular}
\end{table}

\section{Conclusion and future work}
\label{sec:conclusions}  
Optimizing the total loss using the 0-1 loss function leads to the design of an efficient classification algorithm that finds the optimal separator hyperplane and is robust against outliers. This problem is, however, NP-Hard, making it an impractical approach for real-world problems. In addition, the 0-1 loss function does not provide any information about the empirical risk associated with the found hyperplane and is sensitive to imbalances in the number of instances in different classes. We proposed an efficient procedure to optimize the total loss of a modified 0-1 loss function in a one-dimensional case. We extended this approach using an evolutionary algorithm to solve multi-dimensional binary and multi-class classification problems. We designed an objective function for our method to ensure the separator hyperplane has an optimal margin (minimum empirical risk) from instances in each class, improving generalization ability for the classification of unseen instances. We also added the $ L_1 $ regularization term to the algorithm to enable its use for feature selection. We then compared the proposed algorithm, the Optimal-margin Evolutionary Classifier (OEC), on 19 standard classification problems against six state-of-the-art classification algorithms. Results showed that, on the tested datasets, OEC is unbeaten over training sets, even in comparison to the most recent algorithms that provably find a near-optimal separator hyperplane. The generalization ability of the algorithm was also better than other algorithms in our tests, indicating that OEC reliably classifies unseen instances. We also showed that the modified 0-1 loss function is insensitive to class imbalance. To the best of our knowledge, OEC is the first algorithm that optimizes the 0-1 loss function directly and considers the empirical risk (maximum margin), and yet has a running time that is practical for real-world applications. These results are exciting as they encourage the use of evolutionary algorithms for classification, a task that is often, in practice, on-demand. 

One potential next step is to add non-linearity to the algorithm to deal with non-linear classification. One possibility is to incorporate non-linear functions frequently used in Perceptrons \cite{haykin2004comprehensive}. The running time of the algorithm can also be improved by incorporating smart initialization, i.e., hybridization with SVM or LDA. Bagging and boosting \cite{zhou2012ensemble} are other possibilities to improve the generalization ability of the proposed OEC. Graphical Processing Unit (GPU) implementation of OEC may also be another promising direction to ultimately yield an effective evolutionary-based classifier for real-world problems.

\section*{Appendix}
Table \ref{Tbl:comparisonResultsDisc-full} shows the full comparison results between OEC and other classification methods on binary classification problems.
\begin{table}[]
	\centering
	\caption{The values in each row are the average values (over 100 runs). The row "Time" denotes the average time in milliseconds over 100 runs. For some cases, the time has been reported in seconds for which the time values has been post-fixed by the character 's'.}
	\label{Tbl:comparisonResultsDisc-full}
	\begin{tabular}{|l|l|l|l|l|l|l|}
		\hline
		\textbf{Dataset} & \textbf{DEL} & \textbf{WSV} & \textbf{SMP} & \textbf{SVM} & \textbf{MBA} & \textbf{OEC} \\ \hline
		BC-Time & 15 & 0.6 & 112.5 & 13.9 & 6.9 & 78.2  \\\hline
		BC-Train & 96.89* & 94.18* & 97.45* & 96.88* & 95.34* & \textbf{98.42}  \\ \hline
		BC-Test & 96.48* & 93.67* & \textbf{97.24}- & 96.21* & 94.86* & 96.99  \\ \hline
		CG-Time & 10.7 & 0.3 & 98.7 & 5.5 & 4.1 & 38.8  \\\hline
		CG-Train & 94.6* & 96.49* & 95.39* & 95.17* & 95.43* & \textbf{98.37}  \\\hline
		CG-Test & 94.57* & \textbf{96.52}+ & 95* & 95.32- & 95.38- & 95.25  \\ \hline
		GC-Time & 11.2 & 0.4 & 105.2 & 5.1 & 4.1 & 51  \\ \hline 
		GC-Train & 92.33* & 88.26* & 93.47* & 92.4* & 89.48* & \textbf{96.99}  \\\hline
		GC-Test & 89.68- & 85.56* & 90.09- & 87* & 86.47* & \textbf{90.3}  \\ \hline
		PR-Time & 14.3 & 0.4 & 121.2 & 5.9 & 4.3 & 101.3  \\ \hline
		PR-Train & 78.95* & 88.42* & 85.58* & 82.23* & 85.07* & \textbf{92.44}  \\ \hline 
		PR-Test & 74.52* & \textbf{80.14}- & 77.85* & 77.11* & 79.28- & 79.23  \\ \hline
		IS-Time & 19.6 & 0.5 & 131.6 & 11.9 & 4.7 & 150.1  \\ \hline
		IS-Train & 87.21* & 70.54* & 89.79* & 94.2* & 88.58* & \textbf{95.15}  \\ \hline
		IS-Test & 81.7- & 68.05* & 83.2+ & \textbf{84.59}+ & 81.78- & 82.42  \\ \hline
		PD-Time & 14.9 & 0.5 & 99.8 & 22.7 & 4.4 & 72.5  \\ \hline
		PD-Train & 75.57* & 74.6* & 75.63* & 73.09* & 73.03* & \textbf{79.28}  \\ \hline 
		PD-Test & \textbf{74.57}+ & 73.97- & 73.82* & 72.34* & 72.33* & 73.11  \\ \hline
		GR-Time & 26.9 & 0.6 & 107 & 106.3 & 5.3 & 151.7  \\ \hline 
		GR-Train & 73.25* & 74.08* & 74.31* & 71.17* & 71.32* & \textbf{79.8}  \\ \hline 
		GR-Test & 70.87- & 71.71+ & 71.68+ & 68.53* & 69.09* & 70.73  \\ \hline
		VT-Time & 12.6 & 0.4 & 104.7 & 5.3 & 4.2 & 59.6  \\ \hline
		VT-Train & 91.09* & 92.02* & 91.03* & 91.4* & 91.46* & \textbf{93.66}  \\ \hline 
		VT-Test & 90.72- & \textbf{92.06}+ & 90.68- & 90.66- & 90.69- & 90.57  \\ \hline
		MD-Time & 2.4s & 43.9 & 4.0\textit{s} & 28.4\textit{s} & 92.9 & 23.1\textit{s}  \\ \hline
		MD-Train & 72.13* & 79.49* & 76.22* & 79.31* & 76.25* & \textbf{81.99}  \\ \hline
		MD-Test & 54.37* & 54.19* & 54.93- & 54.4* & 54.95- & \textbf{55.15}  \\ \hline
		HV-Time & 67.5 & 1.4 & 231 & 155.3 & 8.3 & 1.3\textit{s}  \\ \hline
		HV-Train & 63.53* & 72.9* & 73.04* & 58.67* & 69.1* & \textbf{93.04}  \\ \hline
		HV-Test & 62.65* & 72.22* & 70.84* & 58.47* & 66.73* & \textbf{92.66}  \\ \hline
		HN-Time & 82.9 & 1.8 & 301.8 & 128.8 & 9 & 1.7\textit{s}  \\ \hline
		HN-Train & 63.56* & 75.46* & 73.12* & 61.82* & 71.69* & \textbf{87.81}  \\ \hline
		HN-Test & 61.86* & 72.93* & 69.07* & 60.39* & 67.54* & \textbf{85.15}  \\ \hline
		SR-Time & 1.9\textit{s} & 41.5 & 927.7 & 141.9\textit{s} & 95.6 & 14.5\textit{s}  \\ \hline
		SR-Train & 61.31* & \textbf{74.14}- & 62.24* & 58.02* & 58.96* & 74.53  \\ \hline
		SR-Test & 56.47* & \textbf{72.75}+ & 56.65* & 55.17* & 55.95* & 68.14  \\ \hline
	\end{tabular}
\end{table}

Table \ref{Tbl:comparisonResultsDiscnoise} shows the full comparison results between OEC and other classification methods on binary classification problems with 15\% of noise.
\begin{table}[]
	\centering
	\caption{15\% noise}
	\label{Tbl:comparisonResultsDiscnoise}
	\begin{tabular}{|l|l|l|l|l|l|l|}
		\hline
		Dataset & \textbf{DEL} & \textbf{WSV} & \textbf{SMP} & \textbf{SVM} & \textbf{MBA} & \textbf{OEC} \\ \hline
		BC-Train & 95.35* & 94.58* & 95.48* & 95.23* & 94.47* & \textbf{97.13}  \\ \hline 
		BC-Test & \textbf{96.98}+ & 95.08* & 96.6+ & 95.56* & 94.81* & 96.3  \\ \hline
		CG-Train & 90.25* & 83.57* & 89.14* & 93.24* & 91.96* & \textbf{96.3}  \\ \hline 
		CG-Test & 90.43* & 82.07* & 89.8* & 93.53* & 92.55* & \textbf{94.3}  \\ \hline
		GC-Train & 88.56* & 65.22* & 93.61* & 92.85* & 89.53* & \textbf{96.78}  \\ \hline 
		GC-Test & 84.47* & 56.26* & \textbf{90.35}- & 86.4* & 85.31* & 90.17  \\ \hline
		PR-Train & 77.42* & 80.28* & 85.6* & 84.08* & 83.6* & \textbf{92.04}  \\ \hline 
		PR-Test & 72.57* & 70.43* & \textbf{78.96}+ & 78.14+ & 77.95+ & 76.53  \\ \hline
		IS-Train & 88.74* & 73.43* & 90.21* & 95.13* & 88.32* & \textbf{95.81}  \\ \hline 
		IS-Test & 81.18* & 68.25* & 82.32* & \textbf{83.5}- & 79.28* & 82.86  \\ \hline
		PD-Train & 74.99* & 75.31* & 74.46* & 70.97* & 70.51* & \textbf{78.35}  \\ \hline  
		PD-Test & 74.15- & \textbf{74.7}+ & 73.03* & 70.36* & 70.17* & 73.74  \\ \hline
		GR-Train & 72.29* & 70.98* & 76.49* & 73.99* & 72.54* & \textbf{81.71}  \\ \hline  
		GR-Test & 67.44* & 65.25* & \textbf{71.66}+ & 70.38- & 69.42* & 70.62  \\ \hline
		VT-Train & 90.85* & 91.09* & 91.36* & 91.32* & 91.41* & \textbf{93.77}  \\ \hline  
		VT-Test & 89.99* & 90.14- & 90.04* & 90.18* & 90.41- & \textbf{90.45}  \\ \hline
		MD-Train & 67.49* & 65.4* & 77.12* & 79.28- & 76.98* & \textbf{79.45}  \\ \hline
		MD-Test & 52.99* & 51.18* & 54.89* & 54.72* & 54.73* & \textbf{57.74}  \\ \hline
		HV-Train & 63.15* & 61.31* & 71.69* & 63.36* & 68.39* & \textbf{90.66}  \\ \hline
		HV-Test & 59.79* & 53.71* & 66.53* & 56.66* & 62.57* & \textbf{92.73}  \\ \hline
		HN-Train & 62.44* & 63.44* & 66.31* & 62.98* & 65.86* & \textbf{86.24}  \\ \hline
		HN-Test & 58.56* & 53.98* & 58.74* & 55.19* & 58.01* & \textbf{85.66}  \\ \hline
		SR-Train & 49.91* & 54.98* & 54.31* & 55.08* & 50.56* & \textbf{73.59}  \\ \hline
		SR-Test & 41.45* & 49.81* & 47.11* & 53.65* & 50.38* & \textbf{67.54}  \\ \hline
	\end{tabular}
\end{table} 

Table \ref{Tbl:comparisonResultsDiscMulticlass-full} shows the full comparison results between OEC and other classification methods on multi-class classification problems.
\begin{table}[]
	\centering
	\caption{One-vs-one strategy, no parameter setting as it is very time consuming.}
	\label{Tbl:comparisonResultsDiscMulticlass-full}
	\begin{tabular}{|l|l|l|l|l|l|}
		\hline
		Dataset & \textbf{DEL} & \textbf{WSV} & \textbf{MBA} & \textbf{SVM} & \textbf{OEC} \\ \hline
		IR-Train & 80.53* & 98.09* & 98.33* & 98.23* & \textbf{99.49}  \\ \hline 
		IR-Test &  80.03* & 97.83+ & 98.13+ & 97.7+ & 96.67  \\ \hline
		IW-Train & 94.91* & 99.39* & 99.99- & \textbf{100}- & 100  \\ \hline
		IW-Test & 93.96* & 98.34+ & \textbf{98.78+} & 97.65- & 97.8  \\ \hline
		TF-Train & 71.82* & 66.98* & 94.06* & 83.57* & \textbf{97.85} \\ \hline
		TF-Test & 71.77* & 66.2* & 92.74* & 82.31* & \textbf{95.6}   \\ \hline
		YD-Train & 66.89* & 69.12* & 76.05* & 79.99* & \textbf{83.44}  \\ \hline
		YD-Test & 67.47* & 67.5* & 72.86* & \textbf{78.33}+ & 75.73  \\ \hline
		RQ-Train & 59.99* & 73.06* & 73.76* & 59.96* & \textbf{81.22}  \\ \hline
		RQ-Test & 56.5* & 61.82* & 62.44* & 58.6* & \textbf{64.04}  \\ \hline
		WQ-Train & 57.6* & 68.85* & 63.19* & 56.7* & \textbf{74.91}  \\ \hline
		WQ-Test & 55.37* & 58.41* & 59.88* & 56.21* & \textbf{61.49}  \\ \hline
		HD-Train & 88.98* & 99.66* & 100- & 100- & 100  \\ \hline
		HD-Test & 89.27* & 86.11* & 85.64* & 94.32- & \textbf{94.99}  \\ \hline
	\end{tabular}
\end{table}

\small

\bibliographystyle{IEEEtran}
\bibliography{References}

\end{document}